\documentclass[letterpaper, 10 pt, conference]{ieeeconf}  % Comment this line out if you need a4paper

\IEEEoverridecommandlockouts                              % This command is only needed if 
\usepackage{graphics} % for pdf, bitmapped graphics files
\usepackage{epsfig} % for postscript graphics files
\usepackage{times} % assumes new font selection scheme installed
\usepackage{amsmath} % assumes amsmath package installed
\usepackage{amssymb}  % assumes amsmath package installed
\usepackage{cite}
\usepackage{array}
\usepackage{float}
\usepackage{bm}
\usepackage{multirow}
\usepackage{booktabs}
\usepackage{subfigure}
\usepackage{algorithm}  
\usepackage{algorithmic}
\title{\LARGE \bf
Conditional Generative Neural System for Probabilistic \\Trajectory Prediction
}

\author{Jiachen Li, Hengbo Ma and Masayoshi Tomizuka
	\thanks{J. Li, H. Ma and M. Tomizuka are with the Department of Mechanical Engineering, 
		University of California, Berkeley, CA 94720, USA
		(e-mail: {\tt\small jiachen\_li, hengbo\_ma, tomizuka@berkeley.edu})}}

\begin{document}

\maketitle
\thispagestyle{empty}
\pagestyle{empty}

%%%%%%%%%%%%%%%%%%%%%%%%%%%%%%%%%%%%%%%%%%%%%%%%%%%%%%%%%%%%%%%%%%%%%%%%%%%%%%%%
\begin{abstract}
Effective understanding of the environment and accurate trajectory prediction of surrounding dynamic obstacles are critical for intelligent systems such as autonomous vehicles and wheeled mobile robotics navigating in complex scenarios to achieve safe and high-quality decision making, motion planning and control. Due to the uncertain nature of the future, it is desired to make inference from a probability perspective instead of deterministic prediction.
In this paper, we propose a conditional generative neural system (CGNS) for probabilistic trajectory prediction to approximate the data distribution, with which realistic, feasible and diverse future trajectory hypotheses can be sampled. 
The system combines the strengths of conditional latent space learning and variational divergence minimization, and leverages both static context and interaction information with soft attention mechanisms.
We also propose a regularization method for incorporating soft constraints into deep neural networks with differentiable barrier functions, which can regulate and push the generated samples into the feasible regions.
The proposed system is evaluated on several public benchmark datasets for pedestrian trajectory prediction and a roundabout naturalistic driving dataset collected by ourselves. The experimental results demonstrate that our model achieves better performance than various baseline approaches in terms of prediction accuracy.
\end{abstract}

%%%%%%%%%%%%%%%%%%%%%%%%%%%%%%%%%%%%%%%%%%%%%%%%%%%%%%%%%%%%%%%%%%%%%%%%%%%%%%%%
\section{INTRODUCTION}
It is desired for a multi-agent prediction system to satisfy the following requirements to generate diverse, realistic future trajectories.
1) \textit{Context-aware}: The system should be able to forecast trajectories which are inside the traversable regions and collision-free with static obstacles in the environment. For instance, when the vehicles navigate in a roundabout (see Fig. 1(a)) they need to advance along the curves and avoid collisions with road boundaries.
2) \textit{Interaction-aware}: The system needs to generate reasonable trajectories compliant to traffic or social rules, which takes into account interactions and reactions among multiple entities. For instance, when the vehicles approach an unsignalized intersection (see Fig. 1(b)), they need to anticipate others' possible intentions and motions as well as the influences of their own behaviors on surrounding entities.
3) \textit{Feasibility-aware}: The system should anticipate naturalistic and physically-feasible trajectories which are compliant to vehicle kinematics or dynamics constraints, although these constraints can be ignored for pedestrians due to the large flexibility of their motions.
4) \textit{Probabilistic prediction}: Since the future is full of uncertainty, the system should be able to learn an approximated distribution of future trajectories close to data distribution and generate diverse samples which represent various possible behavior patterns.
\begin{figure}[!tbp]
	\centering
	\includegraphics[width=\columnwidth]{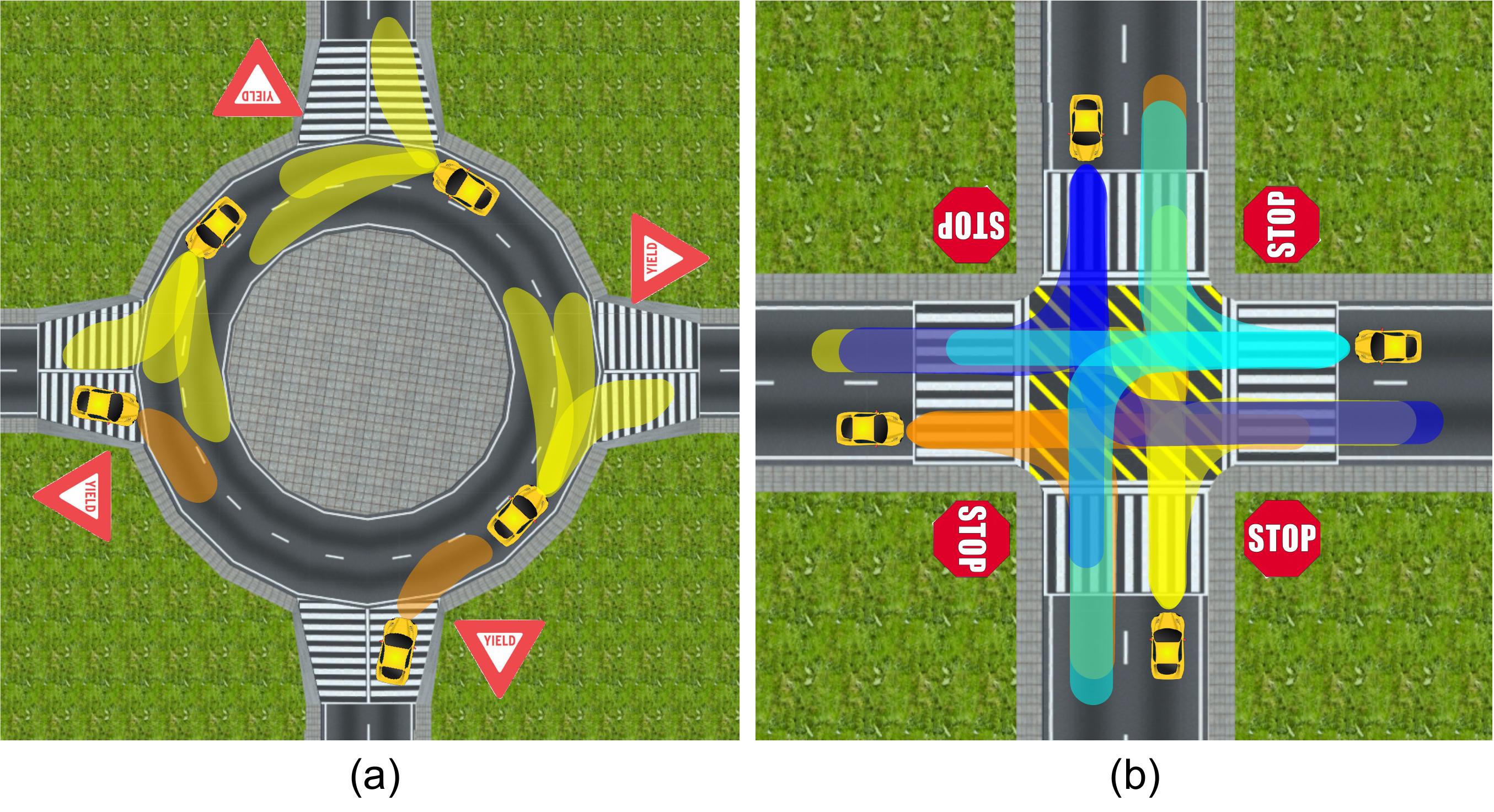}
	\caption{Typical urban traffic scenarios with large uncertainty and interactions among multiple entities. The shaded areas represent the reachable sets of possible trajectories. (a) Unsignalized roundabout with four-way yield signs; (b) Unsignalized intersection with four-way stop signs.}
	%\vspace{1cm}
\end{figure}

In this work, we propose a generative neural system that satisfies all the aforementioned requirements for predicting trajectories in highly interactive scenarios. The system takes advantage of both explicit and implicit density learning in a unified generative system to predict the distributions of trajectories for multiple interactive agents, from which the sampled hypotheses are not only reasonable and feasible but also cover diverse possible motion patterns.

The main contributions of this paper are as follows:
\begin{itemize}
	\item A Conditional Generative Neural System (CGNS) is proposed to jointly predict future trajectories of multiple highly-interactive agents, which takes into account the static context information, interactions among multiple entities and feasibility constraints.
	\item A block attention mechanism and a Gaussian mixture attention mask are proposed and applied to historical trajectories and scene image sequences respectively, which are computationally efficient.
	\item An effective strategy for soft constraint incorporation into deep neural networks is presented.
	\item The latent space learning and variational divergence minimization approaches are integrated into a unified framework in a novel fashion, which combines their strengths on distribution learning. 
	\item The proposed CGNS is validated on multiple pedestrian trajectory forecasting benchmarks and is used to solve a task of anticipating motions of on-road vehicles navigating in highly-interactive scenarios.
\end{itemize}

\section{Related Work}
In this section, we provide a brief overview on related research and illustrate the distinction and advantages of the proposed generative system.
\vspace{0.1cm}

\noindent
\textbf{Trajectory and Sequence Prediction}

Many research efforts have been devoted to predict behaviors and trajectories of pedestrians and on-road vehicles.
Many classical approaches were employed to make time-series prediction, such as variants of Kalman filter based on system process models, time-series analysis and auto-regressive models. 
However, such methods only suffice for short-term prediction in simple scenarios where interactions among entities can be ignored.
More advanced learning-based models have been proposed to cope with more complicated scenarios, such as hidden Markov models \cite{Jiachen_prediction, Wei_benchmark}, Gaussian mixture regression \cite{Jiachen_tracking,Jiachen_ITS}, Gaussian process, dynamic Bayesian networks, and rapidly-exploring random tree. 
However, these approaches are nontrivial to handle high-dimensional data and require hand-designed input features, which confines the flexibility of representation learning. Moreover, these methods only predict behaviors for a certain entity. 
A few works also took advantage of both recurrent neural networks \cite{A9,B1} and generative modeling to learn an explicit or implicit trajectory distribution, which achieved better performance \cite{Jiachen_IV19_1,Jiachen_IV19_2,jiachen_prediction2}. However, they either leveraged only static context images or only trajectories of agents, which is not sufficient to make predictions for the agents that interact with both static and dynamic obstacles.
In this paper, we propose a conditional generative neural system which can leverage both historical scene evolution information and trajectories of multiple interactive agents and generate realistic and diverse trajectory hypotheses.
\vspace{0.1cm}
\begin{figure*}[!tbp]
	\centering
	\includegraphics[width=\textwidth]{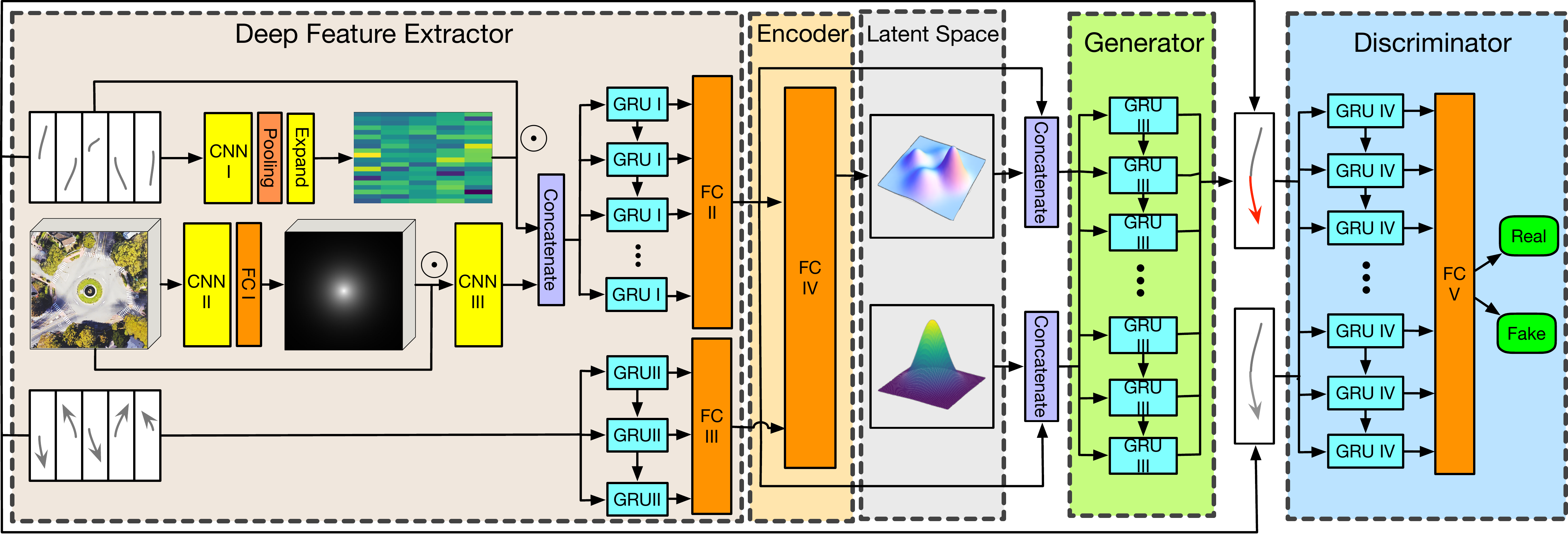}
	\caption{The overview of proposed conditional generative neural system (CGNS), which consists of four key components: (a) A deep feature extractor with soft attention mechanism, which extracts multi-level features from scene context image sequences and trajectories; (b) An encoder to learn conditional latent space representations; (c) A generator (decoder) to sample future trajectory hypotheses; (d) A discriminator to distinguish predicted trajectories from groundtruth.}
\end{figure*}

\noindent
\textbf{Soft Attention Mechanisms}

Soft attention mechanisms have been widely used in neural networks to enable the capability of focusing on a subset of input features, which have been extensively studied in the field of image captioning \cite{attention-2}, visual object tracking \cite{attention-1} and natural language processing. Several works also brought attention mechanisms into trajectory prediction tasks to figure out the most informative and related obstacles \cite{A1,A3,A14,A6}.
In this paper, we put forward a block attention mask mechanism for trajectories to extract the most critical features of each entity as well as a Gaussian mixture attention mechanism for context images to extract the most crucial static features. 
\vspace{0.1cm}

\noindent
\textbf{Deep Bayesian Generative Modeling}

The objective of generative models is to approximate the true data distribution, with which one can generate new samples similar to real data points with a proper variance.
Generative models have been widely employed in tasks of representation learning and distribution approximation in literature, which basically fall into two categories: explicit density models and implicit density models \cite{gan-tutorial}. 
In recent years, since deep neural networks have been leveraged as universal distribution approximators thanks to its high flexibility, two deep generative models have been widely studied: Variational Auto-Encoder (VAE) \cite{VAE} and Generative Adversarial Network (GAN) \cite{gan}. 
Since in trajectory forecasting tasks the predicted trajectories are sampled from the posterior distribution conditioned on historical information, the two models were extended to their conditional versions which results in conditional VAE (CVAE) \cite{A13} and conditional GAN (CGAN) \cite{cgan,A1}.
In this paper, we combine the strengths of conditional latent space learning via CVAE and variational divergence minimization via adversarial training.

\section{Problem Formulation}
The objective of this paper is to develop a deep generative system that can accurately forecast motions and trajectories for multiple agents simultaneously. The system should take into account the historical state information, static context and interactions among dynamic entities.

Assume there are in total $N$ entities in the observation area, which may vary in different cases. 
We denote a set of trajectories covering the history and prediction horizons ($T_h$ and $T_f$) as
\begin{equation}
\resizebox{0.90\hsize}{!}{
	$\mathbf{T}_{k-T_h:k+T_f}=\{\mathbf{t}^i_{k-T_h:k+T_f}|\mathbf{t}^i_{k}=(x^i_k, y^i_k),i=1,...,N\}$}
\end{equation}
where $(x,y)$ is the 2D coordinate in the pixel space or world space.
The latent random variable is denoted as $\mathbf{z}_{k}$, where $k$ is the current time step.
The sequence of context images up to time step $k$ is denoted as $\mathbf{I}_{k-T_h:k}$.
Our goal is to predict the conditional distribution of future trajectories given the historical context images and trajectories $p(\mathbf{T}_{k+1:k+T_f}|\mathbf{T}_{k-T_h:k}, \mathbf{I}_{k-T_h:k})$.
The long-term prediction is realized by propagating the generative system multiple times to the future.
To simplify the notations in the following sections, we denote the condition variable as $C = \{\mathbf{T}_{k-T_h:k},\mathbf{I}_{k-T_h:k}\}$, the sequence of predicted variables as $Y = \{\mathbf{T}_{k+1:k+T_f}\}$.

\section{Methodology}
In this section, we first provide an overview of the key components and the architecture of the proposed Conditional Generative Neural System (CGNS). The detailed theories and models of each component are then illustrated.

\subsection{System Overview}
The architecture of CGNS is shown in Fig. 2 where there is a deep feature extractor (DFE) with an environment attention mechanism (EAM) as well as a generative neural sampler (GNS). 
First, the DFE extracts deep features from a sequence of historical context images and trajectories of multiple interactive agents to obtain both the information of static and dynamic obstacles, where the EAM tells which areas and dynamic entities should be paid more attention to than others when predicting the trajectory of a certain entity.
The above information is utilized as the input of GNS which takes advantage of a deep latent variable model and a variational divergence minimization approach to generate a set of feasible, realistic and diverse future trajectories of all the involved entities.
All the components are implemented with deep neural networks thus can be trained end-to-end efficiently and consistently.

\subsection{Environment-Aware Deep Feature Extraction}
We take advantage of both context images and historical trajectories of interactive agents to extract deep features of both static and dynamic environments. 
In order to figure out the most crucial parts to consider when forecasting behaviors of certain agents, we propose a soft block attention mechanism applied to trajectories and a Gaussian mixture attention mechanism applied to context images. The details are illustrated below.

The historical and future trajectories are constructed as matrices which are treated as 2D images. The former is fed into a convolutional neural network (CNN) and an average pooling layer to obtain a contractable attention mask over the whole trajectory matrix, which is then expanded to the same size as the trajectory matrix by duplicating each column twice corresponding to coordinates $x$ and $y$. 
The original trajectory matrix is multiplied by the block attention mask elementwisely. This mechanism is not applied to the future trajectory matrix since it is unreasonable to have particular attention on the future evolution.
The context image sequences are also fed into a CNN followed by fully connected layers to obtain a set of parameters of the Gaussian mixture distribution, which is used to calculate the context attention mask. The elementwise multiplication of original images and attention masks is fed to a pre-trained feature extractor, which is the convolution base of VGG-19 \cite{vgg19} in this paper. The interaction-aware features and context-aware features are concatenated and fed into a recurrent layer followed by fully connected layers to obtain a comprehensive and consistent feature embedding.

\subsection{Deep Generative Sampling}
The GNS is composed of an encoder $E$ and a generator $G$. The goal of encoder is to learn a consistent distribution in a lower-dimensional latent space, from which the latent variable can be sampled efficiently. The generator aims to produce trajectories as real as possible. An auxiliary discriminator $D$ is adopted, which aims to distinguish fake trajectories from groundtruth.
The generator $G$ and discriminator $D$ formulates a minimax game. The three components can be optimized jointly via conditional latent space learning and variational divergence minimization.
\vspace{0.1cm}

\noindent
\textbf{Conditional Latent Space Learning (CLSL)}

The conditional latent variable model defined in this paper contains three classes of variables: condition variable $C$, predicted variable $Y$ and latent variable $z$. We aim to obtain the conditional distribution $p(Y|C)$.
Given the training data $(C,Y)$, the model first samples $z$ from an arbitrary distribution $Q$. 
Our goal is to maximize the variational lower bound, which is written as
\begin{equation}
\resizebox{0.85\hsize}{!}{
	$\begin{aligned}
	&\log p(Y|C) - \mathcal{D}_{KL}[Q(z|C,Y)||p(z|C,Y)] = \\
	&\mathbb{E}_{z \sim Q} [\log p(Y|z,C)] - \mathcal{D}_{KL}[Q(z|C,Y)||p(z|C)].
	\end{aligned}$}
\end{equation}
where $p(z|C) = \mathcal{N}(0,I)$.
This process can be realized with a Conditional Variational Auto-Encoder which consists of an encoder network $E$ to obtain $Q(z|C,Y)$ and a decoder (generator) network $G$ to model $p(Y|z,C)$.
The loss function can be formulated as a weighted sum of the reconstruction error and KL divergence:
\begin{equation}
\small
\resizebox{0.90\hsize}{!}{
	$\mathcal{L}^{G,E}_{\text{RC}} = \mathop{\mathbb{E}}_{\mathbf{t}_{k+1:k+T_f}, \mathbf{z}_{k} \sim Q}  \left[ \|\mathbf{t}_{k+1:k+T_f} - G(C_k, \mathbf{z}_{k})\|^2 \right]$,}
\end{equation}
\begin{equation}
\small
\resizebox{0.70\hsize}{!}{
	$\mathcal{L}^E_{\text{KL}} = \mathop{\mathbb{E}}_{\mathbf{t}_{k+1:k+T_f}} \left[\mathcal{D}_{\text{KL}}(E(C_k)||p(\mathbf{z}_{k})) \right],$}
\end{equation}
where $\mathbf{z}_k \sim \mathcal{N}(0,I)$.
The optimal encoder and generator can be obtained by
\begin{equation}
\small
\begin{split}
G^*, E^* = \ \arg \min_{G,E} \lambda_1\mathcal{L}^{G,E}_{RC} + \lambda_{2}\mathcal{L}^E_{\text{KL}}.
\end{split}
\end{equation}

\noindent
\textbf{Variational Divergence Minimization (VDM)}

Given two conditional distributions $P_{\text{data}}(Y|C)$ and $P_{\text{GNS}}(Y|C)$ with absolutely continuous density function $p_{\text{data}}(Y|C)$ and $p_{\text{GNS}}(Y|C)$ which denotes the real data distribution and its approximation with GNS, the $f$-divergence \cite{f-divergence} is defined as
\begin{equation}
\resizebox{0.85\hsize}{!}{
	$\mathcal{D}_f (P_{\text{data}} \ || \ P_{\text{GNS}}) = \int_{\mathcal{Y}} p_{\text{GNS}}(Y|C) f\left(\frac{p_{\text{data}}(Y|C)}{p_{\text{GNS}}(Y|C)}\right) \text{d}Y$},
\end{equation}
where $f:\mathbb{R}_+ \to \mathbb{R}$ is a convex and lower-semicontinuous function with $f(1)=0$. 
A lower bound of $f$-divergence can be derived with the convex conjugate function $f^*$
\begin{equation}\label{VDM}
\resizebox{0.9\columnwidth}{!}{
	$\begin{split}
	& \quad \mathcal{D}_f(P_{\text{data}} \ || \ P_{\text{GNS}}) \qquad\qquad\qquad\qquad\qquad\\ 
	&\geq  \sup_{T\in\mathcal{T}} \left(\int_{\mathcal{Y}}p_{\text{data}}(Y|C)T(Y|C))\text{d}Y -  \int_{\mathcal{Y}}p_{\text{GNS}}(Y|C)f^*(T(Y|C))\text{d}Y\right) \\
	&= \sup_{T\in\mathcal{T}}(\mathbb{E}_{Y\sim P_{\text{data}}}[T(Y|C)] - \mathbb{E}_{Y\sim P_{\text{GNS}}}[f^*(T(Y|C))]),
	\end{split}$}
\end{equation}
where $\mathcal{T}$ is an arbitrary class of mapping $T:\mathcal{Y}\to\mathbb{R}$.
In order to minimize the variational lower bound in (\ref{VDM}), we can formulate a minimax game of $p_{\text{GNS}}(Y|C)$ and $T(Y|C)$, which are parameterized by $\theta$ and $\phi$, respectively. Then the optimal $\theta^*$ and $\phi^*$ can be obtained by 
\begin{equation}
\resizebox{0.75\hsize}{!}{
	$\begin{aligned}
	\theta^*, \phi^*  =  \arg &\min_{\theta} \max_{\phi} \ \mathbb{E}_{Y \sim p_{\text{data}}(Y|C)}[T_\phi(Y|C)] \\ \ - \ &\mathbb{E}_{Y \sim p_\theta(Y|C)}[f^*(T_\phi(Y|C))].
	\end{aligned}$}
\end{equation}
In this work, we propose to minimize the Pearson-$\chi^2$ divergence between $P_{\text{data}}\!+\!P_{\text{GNS}}$ and $2P_{\text{GNS}}$
\begin{equation}\label{Pearson}
\small
\mathcal{D}_{\chi^2_{\text{Pearson}}} = \int_{\mathcal{Y}} \frac{\left(2p_{\text{GNS}} - (p_{\text{data}}+p_{\text{GNS}})\right)^2}{p_{\text{data}}+p_{\text{GNS}}} \text{d}Y.
\end{equation}
Since (\ref{Pearson}) is intractable, we leverage the adversarial learning techniques with a generator $G$ and a discriminator $D$ implemented as deep networks. 
The adversarial loss functions are derived as 
\begin{equation}
\small
\mathcal{L}^G_{\text{VDM}} = \frac{1}{2}\mathbb{E}_{z_k \sim p(z)}[(D(G(C_k,z_k)))^2],
\end{equation}
\begin{equation}
\small
\begin{aligned}
\mathcal{L}^D_{\text{VDM}} = &\frac{1}{2}\mathbb{E}_{\mathbf{t}_{k+1:k+T_f}}[(D(\mathbf{t}_{k+1:k+T_f})-1)^2] \\+ &\frac{1}{2}\mathbb{E}_{z_k \sim p(z)}[(D(G(C_k,z_k))+1)^2],
\end{aligned}
\end{equation}
To discriminate the effect of latent space learning, we also involve two additional terms $\mathcal{L}^{G,E}_{\text{VDM}}$ and $\mathcal{L}^{D,E}_{\text{VDM}}$ where the input $z_k$ are sampled from the encoded latent distribution.
Thus, the optimal encoder, generator and discriminator by variational divergence minimization can be obtained as
\begin{equation}
\small
\begin{aligned}
&E^*, G^*,D^* = \\&\arg \min_{G,E} \max_{D} \lambda_3(\mathcal{L}^G_{\text{VDM}}+\mathcal{L}^D_{\text{VDM}}) + \lambda_4(\mathcal{L}^{G,E}_{\text{VDM}}+\mathcal{L}^{D,E}_{\text{VDM}}).
\end{aligned}
\end{equation}

\subsection{Soft Constraint Incorporation}
In order to make generated samples compliant to feasibility constraints of vehicle kinematics, we propose to incorporate a differentiable barrier (indicator) function $I(\cdot)$ in the loss function, which enables soft constraints in deep neural networks via pushing predicted trajectories to the feasible regions.  
In this work, we denote the empirical upper bounds on the absolute values of accelerations $\mathbf{a}_{k+1:k+T_f}$ and path curvatures $\mathbf{\kappa}_{k+1:k+T_f}$ as $a_\text{max}$ and $\kappa_\text{max}$, respectively. Then the feasibility loss can be calculated as
\begin{equation}
\resizebox{0.85\hsize}{!}{
	$\begin{aligned}
	\mathcal{L}^{G,E}_{\text{F}} = \ &\alpha_1\mathbb{E}_{\mathbf{a}_{k+1:k+T_f}} \left[\sum_{t=k+1}^{k+T_f}\max \left(0, 
	\text{sgn}(|\mathbf{a}_t| - \mathbf{a}_{\max} \right)) \right] \\
	+ \ &\alpha_2\mathbb{E}_{\mathbf{\kappa}_{k+1:k+T_f}} \left[\sum_{t=k+1}^{k+T_f}\max \left(0, 
	\text{sgn}(\mathbf{|\kappa}_t| - \mathbf{\kappa}_{\max} \right)) \right],
	\end{aligned}$}
\end{equation}
where sgn($\cdot$) refers to the sign function and $\mathbf{a}_t$,$\mathbf{\kappa}_t$ can be calculated with the predicted waypoints.
This loss term is not applied to human trajectory prediction.

\subsection{Conditional Generative Neural System (CGNS)}
We leverage both CLSL and VDM in the proposed system, which provides complementary strengths.
The objective function of the whole system is formulated as
\begin{equation}
\small
\begin{aligned}
\mathcal{L}_{\text{CGNS}} = \lambda_1\mathcal{L}^{G,E}_{\text{RC}} + \lambda_{2}\mathcal{L}^E_{\text{KL}} +
\lambda_3(\mathcal{L}^G_{\text{VDM}}+\mathcal{L}^D_{\text{VDM}})\\ + \lambda_4(\mathcal{L}^{G,E}_{\text{VDM}}+\mathcal{L}^{D,E}_{\text{VDM}})+ \lambda_5 \mathcal{L}^{G,E}_{\text{F}},
\end{aligned}
\end{equation}
which can be trained end-to-end.
In practice, due to the existence of reconstruction loss, the generator tends to improve faster than the discriminator, which may result in unbalanced training. Therefore, we compensate the unbalance by training the discriminator multiple times in each iteration.

\section{Experiments}
In this section, we validate the proposed CGNS on three benchmark datasets for trajectory prediction which are available online and solve a task of probabilistic behavior prediction for multiple interactive on-road vehicles in a roundabout scenario. The model performance is compared with several state-of-the-art baselines.

\subsection{Datasets}

\vspace{0.1cm}
\noindent
\textbf{ETH} \cite{ETH_dataset} \textbf{and UCY} \cite{UCY_dataset}: These datasets include bird-eye-view videos and image annotations of pedestrians in various outdoor and indoor scenarios. The trajectories were extracted in the world space.

\vspace{0.1cm}
\noindent
\textbf{Stanford Drone Dataset (SDD)} \cite{A10}: The dataset also contains a set of bird-eye-view videos and the corresponding trajectories of involved entities, which was collected in multiple scenarios within a university campus full of pedestrians, bikers and vehicles. The trajectories were extracted in the pixel space instead of the world space.

\vspace{0.1cm}
\noindent
\textbf{INTERACTION Dataset (ID)} \cite{Wei_IROS19,Wei_INTERACTION_19}: The raw dataset was collected by a drone with camera and our testing vehicle equipped with LiDAR. The trajectories were extracted by visual detection. We visualized the real trajectories in our simulator to obtain the bird-eye-view images, where the static context information came from the Google Earth. 

\subsection{Evaluation Metrics and Baselines}
We evaluate the model performance in terms of average displacement error (ADE) defined as the average distance between the predicted trajectories and the groundtruth over all the involved entities within the prediction horizon, as well as final displacement error (FDE) defined as the distance at the last predicted time step.
To allow for fair comparisons with prior works \cite{A1,A8,A9}, we predicted the future 12 time steps (4.8s) based on the previous 8 time steps (3.2s) for ETH and UCY in the Euclidean space. We used the standard training and testing split for SDD and make predictions in the pixel space.
For our own dataset ID, we predicted the future 10 time steps (5s) based on the historical 4 time steps (2s) in the Euclidean space.
\begin{table*}[!tbp]
	\small
	\label{tab:feature}
	\caption{ADE / FDE Comparisons of Pedestrian Trajectory Prediction (ETH and UCY dataset).}
	\vspace{-0.5cm}
	\begin{center}
		\begin{tabular}{m{1.2cm}<{\centering}| m{1.5cm}<{\centering}| m{1.5cm}<{\centering}| m{1.5cm}<{\centering}| m{1.5cm}<{\centering}| m{1.5cm}<{\centering}| m{1.5cm}<{\centering}| m{1.5cm}<{\centering}| m{1.5cm}<{\centering} }
			\toprule
			\midrule
			&  CVM & LR & P-LSTM & S-LSTM & S-GAN & S-GAN-P  & SoPhie  & \textbf{CGNS}\\ % \hhline{=|=|=}
			\midrule 
			ETH   & 1.42 / 2.88 &  1.33 / 2.94 & 1.13 / 2.38 & 1.09 / 2.35 & 0.81 / 1.52 & 0.87 / 1.62 & 0.70 / 1.43 & \textbf{0.62} / \textbf{1.40} \\ 
			HOTEL & 0.51 / 0.68 &  \textbf{0.39} / 0.72 & 0.91 / 1.89 & 0.79 / 1.76 & 0.72 / \textbf{0.61} & 0.67 / 1.37 & 0.76 / 1.67 & 0.70 / 0.93 \\ 
			UNIV  & 0.73 / 1.63 &  0.82 / 1.59 & 0.63 / 1.36 & 0.67 / 1.40 & 0.60 / 1.26 & 0.76 / 1.52 & 0.54 / 1.24 & \textbf{0.48} / \textbf{1.22}\\ 
			ZARA1 & 0.59 / 1.36 &  0.62 / 1.21 & 0.44 / 0.84 & 0.47 / 1,00 & 0.34 / 0.69 & 0.35 / 0.68 & \textbf{0.30} / 0.63 & 0.32 / \textbf{0.59} \\
			ZARA2 & 0.84 / 1.55 &  0.77 / 1.48 & 0.51 / 1.16 & 0.56 / 1.17 & 0.42 / 0.84 & 0.42 / 0.84 & 0.38 / 0.78 & \textbf{0.35} / \textbf{0.71} \\
			\midrule
			AVG   & 0.82 / 1.62 &  0.79 / 1.59 & 0.72 / 1.53 & 0.72 / 1.54 & 0.58 / 1.18 & 0.61 / 1.21 & 0.54 / 1.15 & \textbf{0.49} / \textbf{0.97} \\   
			\bottomrule
		\end{tabular}
	\end{center}
	\vspace{-0.4cm}
\end{table*}
\begin{table*}[!tbp]
	\small
	\label{tab:feature}
	\caption{ADE / FDE Comparisons of Pedestrian Trajectory Prediction (SDD dataset).}
	\vspace{-0.5cm}
	\begin{center}
		\begin{tabular}{m{1.2cm}<{\centering}| m{1.5cm}<{\centering}| m{1.5cm}<{\centering}| m{1.5cm}<{\centering}| m{1.5cm}<{\centering}| m{1.5cm}<{\centering}| m{1.5cm}<{\centering}| m{1.5cm}<{\centering}| m{1.5cm}<{\centering} }
			\toprule
			\midrule
			& LR & P-LSTM & S-LSTM & S-GAN  & SoPhie & CAR-Net & DESIRE & \textbf{CGNS}\\ % \hhline{=|=|=}
			\midrule
			SDD     & 37.1 / 63.5 & 35.8 / 55.4 & 31.2 / 57.0 & 24.8 / 38.6 & 17.8 / 32.1 & 25.7 / 51.8 & 19.3 / 34.1 & \textbf{15.6} / \textbf{28.2}\\   
			\bottomrule
		\end{tabular}
	\end{center}
	\vspace{-0.4cm}
\end{table*}
\vspace*{-0.5cm}
\begin{table*}[!tbp]
	\small
	\label{tab:feature}
	\caption{ADE / FDE Comparisons of Vehicle Trajectory Prediction (RD dataset). $\mathbf{T}$ represents only using interaction-aware (trajectory) features and $\mathbf{T}$ + $\mathbf{I}$ represents using additional context-aware (image) features.}
	\vspace{-0.5cm}
	\begin{center}
		\begin{tabular}{m{0.5cm}<{\centering}| m{1.4cm}<{\centering}| m{1.4cm}<{\centering}| m{1.4cm}<{\centering}| m{1.4cm}<{\centering}| m{1.4cm}<{\centering}|  m{1.4cm}<{\centering} | m{1.4cm}<{\centering} | m{1.5cm}<{\centering} | m{1.5cm}<{\centering} }
			\toprule
			\midrule
			\multirow{2}*{\shortstack[lb]{}} 
			& \multicolumn{5}{c|}{Baseline Models} & \multicolumn{4}{c}{Proposed CGNS} \\
			%\midrule
			\cline{2-10}
			& CVM & LR & P-LSTM & S-LSTM & S-GAN  &  $\mathbf{T}$ + CLSL & $\mathbf{T}$ + VDM & $\mathbf{T}$ + CLSL + VDM & $\mathbf{T}$ + $\mathbf{I}$ + CLSL+VDM\\ % \hhline{=|=|=}
			\midrule
			1.0s & \textbf{0.16} / 0.29 & 0.24 / 0.32 & 0.23 / 0.28 & 0.24 / 0.30 & 0.22 / 0.28 & 0.19 / \textbf{0.23} & 0.22 / 0.27 & 0.17 / 0.25  & 0.21 / 0.26 \\  
			2.0s & 0.59 / 0.78 & 0.58 / 0.92 & 0.47 / 0.60 & 0.45 / 0.57 & 0.42 / 0.58 & \textbf{0.34} / 0.42 & 0.38 / 0.45 & 0.38 / 0.44 & 0.35 / \textbf{0.40} \\
			3.0s & 1.21 / 1.92 & 1.43 / 2.28 & 0.84 / 1.53 & 0.80 / 1.48 & 0.81 / 1.54 & 0.72 / 1.33 & 0.75 / 1.37 & 0.69 / 1.24 & \textbf{0.64} / \textbf{1.15} \\
			4.0s & 2.94 / 3.98 & 3.85 / 4.73 & 1.27 / 1.51 & 1.21 / 1.69 & 1.28 / 1.87 & 1.26 / 1.81 & 1.35 / 1.76 & 0.86 / 1.33 & \textbf{0.79} / \textbf{1.23} \\
			5.0s & 4.28 / 6.12 & 5.89 / 6.91 & 1.78 / 2.21 & 1.69 / 2.77 & 1.65 / 2.68 & 1.85 / 3.20 & 1.72 / 2.89 & 1.54 / 2.37 & \textbf{1.47} / \textbf{2.12} \\ 
			\bottomrule
		\end{tabular}
	\end{center}
	\vspace{-0.4cm}
\end{table*}

We compared the performance of our proposed system with the following baseline approaches on multiple datasets: \textit{Constant Velocity Model (CVM)}, \textit{Linear Regression (LR)}, \textit{Probabilistic LSTM (P-LSTM)}, \textit{Social LSTM (S-LSTM)} \cite{A9}, \textit{Social GAN (S-GAN and S-GAN-P)} \cite{A8}, \textit{Clairvoyant attentive recurrent network (CAR-Net)} \cite{A3}, \textit{SoPhie} \cite{A1} and \textit{DESIRE} \cite{A13}.

\subsection{Implementation Details}
Since the whole system consists of differentiable functions approximated by deep neural networks, it can be trained end-to-end efficiently. The detailed model architecture and hyper-parameters are introduced below.

In the deep feature extractor, the $\text{CNN}_\text{I}$ contains one conv-layer with kernel size $5\times5$ and zero padding to keep the same dimension. The $\text{CNN}_\text{II}$ contains three conv-layers with kernel size $3\times3$ and the $\text{FC}_\text{I}$ contains two layers with 64 hidden units. The $\text{CNN}_\text{III}$ is the convolution base of pre-trained VGG-19 whose weights are fixed during training. The $\text{GRU}_\text{I}$, $\text{GRU}_\text{II}$, $\text{FC}_\text{II}$ and $\text{FC}_\text{III}$ all have 128 hidden units.
The Encoder $\text{FC}_\text{IV}$ has three fully-connected layers with 256, 128 and 64 hidden units, respectively. The dimension of encoded latent space is two.
The Generator $\text{GRU}_\text{III}$ has 128 hidden units. The Discriminator $\text{GRU}_\text{IV}$ has 128 hidden units and $\text{FC}_\text{V}$ has three layers with 128, 128 and 1 units, respectively.
In all the experiments, we set $\lambda_1=5.0, \lambda_{2}=\lambda_3=\lambda_4=\lambda_5=1.0$ and $\alpha_1=\alpha_2=1000$.
The Adam optimizer was employed with a learning rate of 0.002. 
Moreover, we found that the Gaussian mixture in the context image attention mask does not lead to obvious improvement in terms of prediction accuracy and diversity than a single Gaussian in this task. Therefore, we utilized the latter to reduce model complexity and show the corresponding results.

\subsection{Quantitative Analysis}
\textbf{ETH and UCY Dataset:} 
To allow for fair comparisons with multiple baseline approaches which only leverage historical trajectory information, we deactivated the branch of context feature extraction in our system to illustrate its superiority on prediction accuracy based on the same input as prior works.
The ADE and FDE of the proposed CGNS and baseline models in Euclidean space are compared in Table I. Some of the reported statistics are adapted from the original papers.

It can be seen that the CVM performs the worst as expected since the constant velocity approximation is insufficient for a crowded scenario with highly interactive agents.
The LR performs slightly better in most scenarios than CVM but achieves the smallest error on the HOTEL dataset. A possible reason is that the human trajectories in this dataset tend to be more straight and smooth, which brings an advantage for linear fitting methods.
The P-LSTM and S-LSTM provide an improvement with similar accuracy due to the exploitation of recurrent neural networks.
The S-GAN and Sophie achieve a bigger progress thanks to the implicit generative modeling of trajectory distribution.
Our approach makes a step forward on prediction accuracy, which implies the effectiveness of latent space learning.

\vspace{0.1cm}
\textbf{Stanford Drone Dataset:}  
We also compared the ADE and FDE of the CGNS and baseline models in pixel space, which is shown in Table II.
Similarly, the linear method LR performs the worst and the ordinary P-LSTM and S-LSTM give a slightly better accuracy.
The CAR-Net makes a step forward by utilizing a physical attention module.
The S-GAN and DESIRE provide better results than the above baselines since they solve the task from a probabilistic perspective by learning implicit data distribution and latent space representations, respectively. 
Our approach achieves the best performance in terms of prediction error, which implies the significance and necessity of leveraging both context and trajectory information.
The combination of CLSL and VDM also contributes to the enhancement.

\vspace{0.1cm}
\textbf{INTERACTION Dataset:}
We finally compared the model performance on our roundabout driving dataset in Table III. The SoPhie, CAR-Net and DESIRE are not involved since their codes are not publicly available.
It is shown that the linear models CVM and LR have similar performance to advanced learning-based models for short-term prediction since the velocity and yaw angle of vehicles cannot vary much in a short period due to kinematics feasibility constraints. However, as the prediction horizon increases, their performance deteriorates much faster. 
A potential reason is that due to the curving roads within the roundabout area, the vehicles tend to advance along the curving lines to avoid collisions, which is not able to be captured by linear approximations.
The P-LSTM and S-LSTM provide similar results, which implies that the social pooling mechanism has little effects on feature extraction in this scenario.
Our CGNS is able to achieve the smallest prediction error among baseline models in most cases especially for long-term prediction.

\subsection{Qualitative Analysis}

We provide a qualitative analysis of the prediction results on our INTERACTION dataset.
To illustrate the effectiveness of the attention module, we visualize the context image masks and trajectory block masks of several typical testing cases in Fig. 3. Detailed analysis can be found in the caption.
The distribution of generated future trajectories is approximated by the kernel density estimation, which is visualized in Fig. 4. 
We can see that the system can generate smooth, feasible and realistic vehicle trajectories, which evolve along the road curves. 
The groundtruth is located at the most dense part of the distribution in most cases.
In general, our proposed CGNS can achieve better generation performance in terms of realism and diversity. 

\subsection{Ablative Analysis}
We conduct an ablative analysis on the RD dataset to demonstrate relative significance of each component in the proposed CGNS. The ADE and FDE of each model setting are shown in Table III.
We notice that using the $\mathbf{T}$ + CLSL and $\mathbf{T}$ + VDM achieves similar performance in terms of prediction error while $\mathbf{T}$ + CLSL + VDM provides a notable improvement. 
Moreover, it is demonstrated that the complete system $\mathbf{T}$ + $\mathbf{I}$ + CLSL + VDM does not lead to obvious improvement compared with three partial systems for short-term prediction while its superiority becomes more remarkable as the forecasting horizon increases. This is reasonable since the static context has little effect on driver behaviors in a short period.
More specifically, since the trajectory segment within a short period can be approximated by a linear segment, learning the road curvature from context images does not provide much assistance for prediction. As the forecasting horizon increases, however, the restriction of road geometry on vehicle motions cannot be ignored any more, which results in larger performance gain of leveraging context information.
\begin{figure*}[!tbp]
	\centering
	\includegraphics[width=\textwidth]{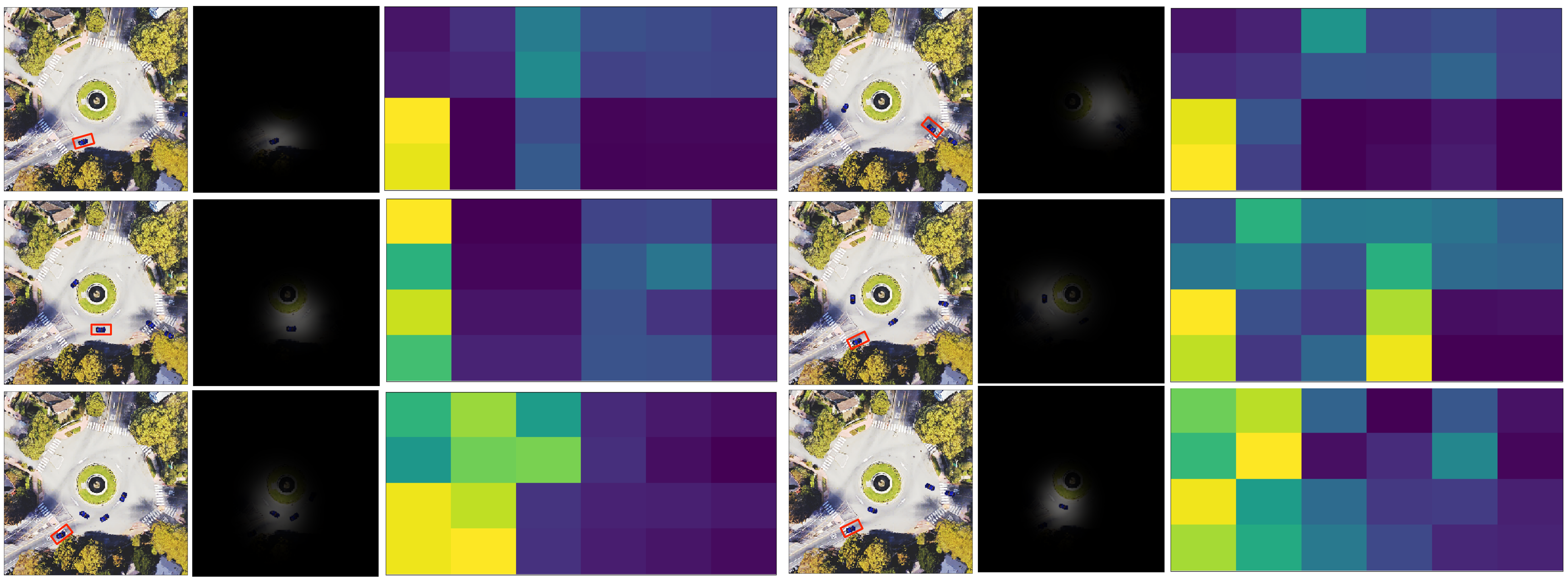}
	\caption{The visualization of the context image masks and trajectory block attention masks. Particularly, in the trajectory masks, there are four rows representing four historical time steps and six columns representing six vehicles in the scene. The first column corresponds to the predicted vehicle and the others corresponds to surrounding ones. Brighter colors indicate larger attention weights. The predicted vehicles are indicated with red bounding boxes. In all the cases, the image masks have a large weight around the predicted vehicle and the area of its heading direction. In the first three cases, only the historical trajectories of the predicted vehicle are assigned large attention weights, which implies that the other vehicles have little effect in these situations. However, in the last three cases, more attention is paid to other vehicles since there exist strong interactions which increases the inter-dependency.}
\end{figure*}
\begin{figure}[!tbp]
	\centering
	\includegraphics[width=\columnwidth]{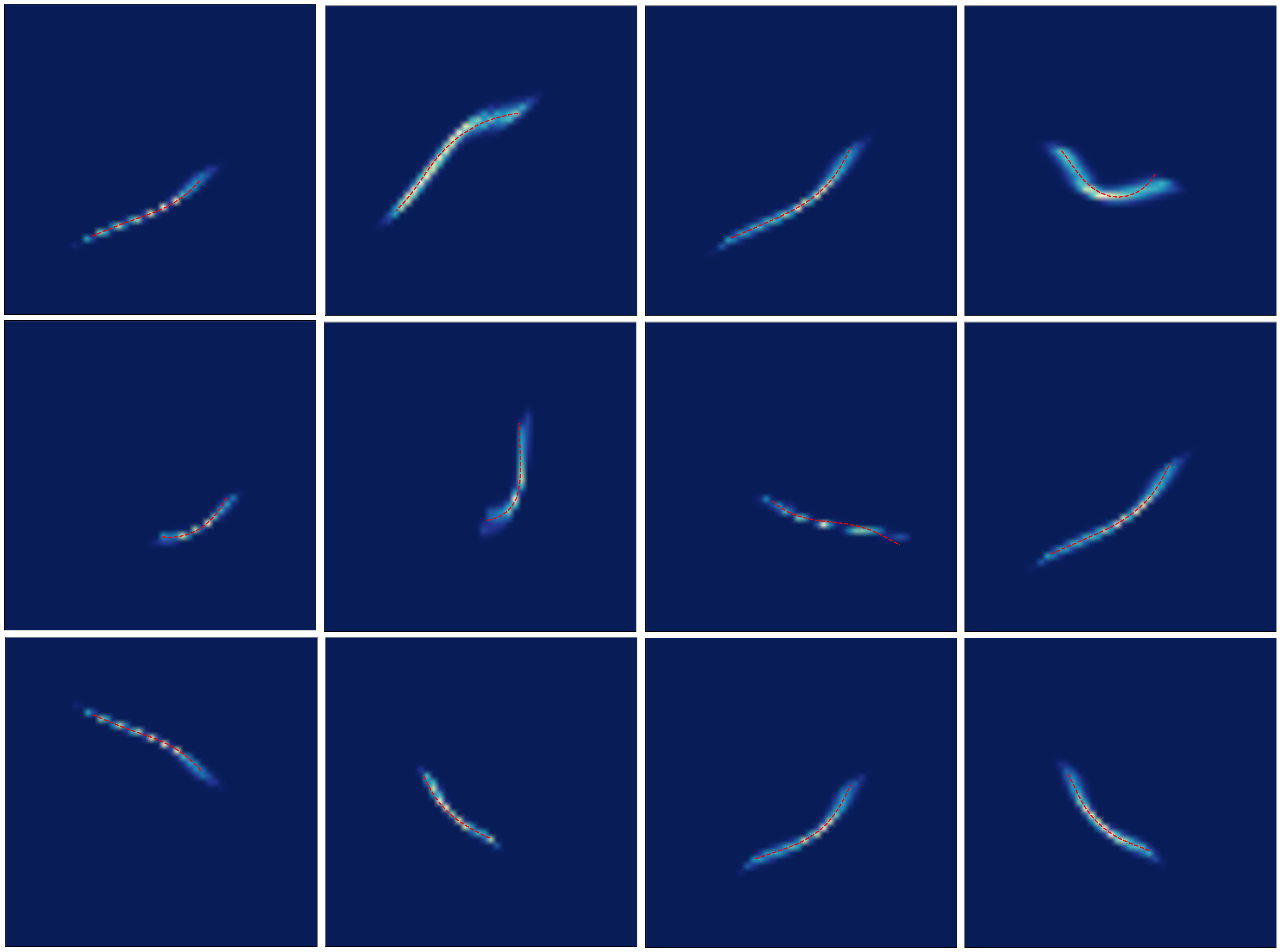}
	\caption{The visualization of sampled future trajectory hypotheses with CLSL+VDM training strategies. The red dashed lines denote groundtruth trajectories.}
\end{figure}

\section{CONCLUSIONS}
In this paper, we propose a conditional generative neural system for long-term trajectory prediction, which takes into account both static context information through images and dynamic evolution of traffic situations through trajectories of interactive agents. We also incorporate attention mechanisms to figure out the most critical portions for predicting motions of a certain entity. The system combines the strengths of both latent space learning and variational divergence minimization to approximate the data distribution, from which realistic and diverse trajectory hypotheses can be sampled. The proposed system is validated on various benchmark datasets as well as a roundabout driving dataset collected by ourselves. The results show that our system can achieve better performance than various baseline models on most datasets in terms of prediction accuracy.

%\addtolength{\textheight}{-12cm}   
%%%%%%%%%%%%%%%%%%%%%%%%%%%%%%%%%%%%%%%%%%%%%%%%%%%%%%%%%%%%%%%%%%%%%%%%%%%%%%%%
%\section*{APPENDIX}


%\section*{ACKNOWLEDGMENT}

%%%%%%%%%%%%%%%%%%%%%%%%%%%%%%%%%%%%%%%%%%%%%%%%%%%%%%%%%%%%%%%%%%%%%%%%%%%%%%%%
%\clearpage
\bibliographystyle{IEEEtran}
\bibliography{reference}

\end{document}